\documentclass[journal]{IEEEtran}

\usepackage{amsmath,amssymb,amsfonts}
\usepackage{newtxtext,newtxmath}
\usepackage{graphicx}
\usepackage{algorithm}
\usepackage{algpseudocode}
\usepackage{booktabs}
\usepackage{array}
\usepackage{xcolor}
\usepackage{url}
\usepackage{cite}
\usepackage[hidelinks]{hyperref}

\newcommand{\R}{\mathbb{R}}
\newcommand{\SEtwo}{\mathrm{SE}(2)}
\newcommand{\Q}{\mathbf{q}}
\newcommand{\X}{\mathbf{x}}
\newcommand{\obs}{\mathcal{O}}
\newcommand{\free}{d_\text{free}}
\DeclareMathOperator*{\argmin}{arg\,min}

\newtheorem{theorem}{Theorem}
\newtheorem{lemma}{Lemma}
\newtheorem{corollary}{Corollary}
\newtheorem{proposition}{Proposition}

\begin{document}

\title{TPS-A*: Path Planning for Any-Shape Vehicles\\ with Certified Collision Checking\\ in Trajectory Parameter Space}

\author{Jose~Luis~Blanco-Claraco,
Shravan~S.~Rai,
Fernando~Cañadas-Aránega,
José~Luis~Torres-Moreno,
and~José~Carlos~Moreno-Úbeda%
\thanks{
(Corresponding author: Jose Luis Blanco-Claraco (e-mail: jlblanco@ual.es).)}%
\thanks{J.L. Blanco-Claraco, J.L. Torres-Moreno are with the Department of Engineering, CIESOL, CeiA3. University of Almería, 04120, Almería, Spain.}%
\thanks{S.S. Rai is with Teladoc Health, Santa Barbara, CA 93117, USA.}%
\thanks{F. Cañadas-Aránega, J.C. Moreno-Úbeda are with Department of Informatics, CIESOL, CeiA3. University of Almería, 04120, Almería, Spain.}%
\thanks{The planner is open source at
\url{https://github.com/MRPT/mrpt_path_planning}, with a ROS~2 wrapper in
\url{https://github.com/mrpt-ros-pkg/mrpt_navigation}. The complete benchmark suite will be released upon publication.}}

\markboth{TPS-A*}%
{Blanco-Claraco \MakeLowercase{\textit{et al.}}: TPS-A*: Path Planning for Any-Shape Vehicles}

\maketitle

\begin{abstract}
Global planners for ground vehicles usually approximate the vehicle by
a disk, trading safety against conservativeness, while exact-footprint
checking pays for every candidate motion, pose, and footprint edge. We present
TPS-A*, a lattice planner whose edges are kinematically feasible
trajectories from parameterized trajectory generators, built on certified
collision checking in trajectory parameter space. A grid precomputed per
footprint returns, in one pass over the local obstacles, a certified free
distance along every candidate trajectory, so the per-node collision
cost is independent of footprint complexity and trajectory count. We prove that
this grid never overestimates the free distance of the
continuously swept, possibly non-convex footprint, and state the planner's
guarantees: collision-free paths and bounded suboptimality relative to its
lattice abstraction. Goals with or without heading are reached exactly by
Reeds-Shepp shots certified as trajectories of the same generators. An exact
independent test finds no collision in its paths. On 300 cluttered worlds, with the true footprint and the same
goal pose for all planners, it solves every query with a collision-free
goal pose in a median of three milliseconds on an AMD Ryzen 7 2700X,
on par with hybrid
A*, 45\% of whose paths overlap obstacles, and nearly sixty times faster than
a state-lattice planner, the only other collision-free one; with
position goals, eighty-nine times faster than
hybrid A*. It matches hybrid A* in automated parking and on narrow doorways, docks
a non-convex fork footprint that no convex model admits, and drove a real robot
across a university campus among people.
\end{abstract}


\section{Introduction}
\IEEEPARstart{G}{lobal} path planners for ground vehicles are commonly built on
a simplified vehicle model: the body is approximated by a disk and the obstacles
are dilated by that disk's radius in a cost
map~\cite{macenski2023survey}. For an elongated or irregularly shaped vehicle
this forces a choice between a circumscribed disk, which rejects passages the
true body fits through, and an inscribed disk, which produces paths the true
body cannot follow without collision. Planners that check the exact polygon
avoid this dilemma, but do so pose by pose, so the cost of every candidate
motion grows with the footprint complexity and with the number of poses sampled
along it~\cite{pan2012fcl,dolgov2010path}.

This paper addresses global path planning for a single ground vehicle of
\emph{arbitrary} planar footprint, possibly non-convex, subject to its
kinematic constraints (minimum turning radius, bounded linear and angular
velocity), among static obstacles. Our goal is a planner whose output is
kinematically feasible, whose collision reasoning is \emph{certified} against
the exact polygon, and whose collision-checking cost does not grow with the
shape of the vehicle.

The approach builds on the Trajectory Parameter Space
(TP-Space)~\cite{blanco2006tpspace,blanco2008extending}. A Parameterized
Trajectory Generator (PTG) is a family of kinematically feasible trajectories
indexed by a selector $k$ and a traveled distance $d$; mapping obstacles into
TP-Space turns the vehicle into a free-flying point. TP-Space was introduced for
reactive navigation~\cite{blanco2006tpspace,blanco2008extending,blanco2008foundations}
and later used by our randomized TP-Space RRT planner~\cite{blanco2015tpsrrt}.

\textbf{Methodological principle.} The central observation of this paper is
that, for lattice search, the TP-Space collision model is not merely a faster
pose checker but a different \emph{interface} between search and collision
checking. A precomputed per-PTG grid stores, for every cell of the
vehicle-local workspace and every trajectory $k$, a lower bound on how far the
vehicle may travel along $k$ before its footprint touches any point in that
cell. One pass of the local obstacles through this grid therefore returns, for
the node being expanded, a certified free distance along \emph{every}
trajectory of the family at once. The search then selects any number of
successors, at any distance below that bound, without further collision
queries. Consequently the per-node collision cost is independent of both the
footprint complexity and the branching factor, whereas pose-by-pose checking
pays for each successor, each sampled pose, and each footprint edge.

\textbf{Contributions.}
\begin{enumerate}
\item The principle above, stated as a \emph{node-batched collision check}
for lattices whose successors are PTG trajectories, with a complexity analysis
and measurements against footprint complexity and branching factor
(Sec.~\ref{sec:collision}, \ref{sec:exp-micro}).
\item A \emph{certified} construction of the TP-Space collision grid, with a
proof that its free distances never exceed the true ones for the continuously
swept, possibly non-convex footprint, a bound on its conservativeness, and an
extension to region obstacles (Theorem~\ref{thm:sound}); the construction
previously used in TP-Space tools is not conservative.
\item TPS-A*, a deterministic $\SEtwo$ lattice planner built on this check,
with a deferred analytic goal expansion and certified Reeds-Shepp shots that
reach full-pose and position goals exactly, and a precise statement of what it
guarantees: certified collision-free, kinematically feasible paths and
$\varepsilon$-bounded suboptimality relative to its lattice
abstraction, for position goals and, with a Reeds-Shepp heuristic, for
full-pose goals, together with what it does not guarantee (strict resolution
completeness) (Sec.~\ref{sec:algorithm}, \ref{sec:theory}).
\item An evaluation on 300 BARN worlds against search-based (Nav2 Hybrid-A*,
SBPL) and sampling-based (OMPL) planners with the true footprint, on narrow
doorways from real floor plans, on the TPCAP automated-parking
benchmark~\cite{li2023tpcap} with a passenger car, on docking with a
non-convex footprint, and on a real Husky robot on a university
campus, with every returned path of
every planner checked by an exact, planner-independent test
(Sec.~\ref{sec:experiments}). No TPS-A* path collides; on the cluttered
worlds it is as fast in median as the fastest baseline, whose paths often
collide, and much faster than the only other planner whose paths never do.
\end{enumerate}

\section{Related Work}\label{sec:related}

\textbf{State-lattice planning.} Search over a lattice of precomputed motion
primitives is a mature
paradigm~\cite{pivtoraiko2005generating,pivtoraiko2009differentially},
realized in SBPL, which runs ARA*~\cite{likhachev2003ara} over an
$(x,y,\theta)$ lattice, and in Nav2's
\texttt{SmacPlannerLattice}~\cite{macenski2023survey}. Primitives end on
lattice states, so the graph is fixed and the classical A* guarantees apply;
SBPL precomputes the cells each primitive sweeps~\cite{likhachev2009planning},
so a primitive costs one cost-map lookup per swept cell. TPS-A* instead
generates successors from PTGs and replaces per-primitive checking by one
TP-Space query per node.

\textbf{Hybrid-A*.} Hybrid-A*~\cite{dolgov2010path}, realized in Nav2 as
\texttt{SmacPlannerHybrid}~\cite{macenski2023survey}, expands continuous states
along Dubins~\cite{dubins1957curves} or Reeds-Shepp~\cite{reeds1990optimal}
arcs, keeps one representative per lattice cell, trading strict resolution
completeness for continuous paths, and ends with analytic curve shots to the
goal. TPS-A* adopts the representatives and the shots, the latter executed and
certified as PTG trajectories (and inherits the completeness caveat,
Sec.~\ref{sec:theory}), but differs in the motion model, any PTG family with
reverse motion, and in the collision model.

\textbf{Sampling-based planning and certified checking.} RRT, RRT*, and their
kinodynamic variants plan by sampling and
steering~\cite{lavalle2001randomized,karaman2011sampling}; arbitrary
footprints are handled by a geometry library such as FCL~\cite{pan2012fcl}
inside OMPL~\cite{sucan2012ompl}, one of our baselines. For articulated robots,
distance bounds between checked configurations certify continuous
motions~\cite{schwarzer2005adaptive}; our soundness argument uses the same
displacement bound, but offline, once per footprint, so every query reuses it.

\textbf{Precomputed collision checking.} Precomputing how the footprint meets a
grid takes the polygon out of the inner loop. Besides swept-cell lists, the body
can be covered by disks, each checked by one lookup in a dilated
grid~\cite{ziegler2010fast}, which is conservative by the covering, must be
recomputed when the map changes, and checks poses only. Closest to our
principle, motion-planning hardware for manipulators precomputes the voxels
swept by every edge of a fixed roadmap and streams the obstacle voxels once to
invalidate all colliding edges~\cite{murray2016robot}. TPS-A* uses this
obstacle-to-motion lookup in a lattice search: since PTG trajectories are
defined in the vehicle frame, one table serves every node after a rigid
transform of the local obstacles.

\textbf{TP-Space.} PTGs and TP-Space collision grids were introduced for
reactive navigation~\cite{blanco2006tpspace,blanco2008extending}, where the
obstacles are mapped for all trajectories at once but only from the current
pose, and reused by TP-Space RRT~\cite{blanco2015tpsrrt}, where each extension
queries one trajectory. New here are this all-trajectory query as the
successor interface of a lattice search, and a provably conservative grid;
Table~\ref{tab:related} summarizes the differences.

\begin{table*}[t]
\centering
\caption{Collision model and guarantees of related planners. $V$: footprint
vertices; $B$: successors per node; $n_p$: poses checked per successor; $S$,
$P$: cells swept by a primitive, on the footprint outline; $N$: local obstacle
points (or voxels); $\bar L$, $\bar E$: mean trajectories per grid cell, roadmap
edges per voxel.}
\label{tab:related}
\small
\setlength{\tabcolsep}{4pt}
\begin{tabular}{llll}
\toprule
Planner & Collision model & Cost per node & Guarantees \\
\midrule
State lattice (SBPL) & swept cells per primitive & $O(B\,S)$ & complete on a fixed graph, $\varepsilon$-suboptimal \\
Hybrid-A* (Nav2 Smac) & outline rasterized per pose & $O(B\,P)$ & representative pruning \\
RRT*/BIT* + FCL (OMPL) & exact polygon, broadphase & $O(B\,n_p\,(V + \log N))$ & probabilistic, anytime \\
Swept-voxel roadmap~\cite{murray2016robot} & voxels per roadmap edge & $O(N\,\bar E)$, all edges & those of the fixed roadmap \\
TP-Space RRT~\cite{blanco2015tpsrrt} & TP-Space grid, per trajectory & $O(N)$ per trajectory & probabilistic \\
\textbf{TPS-A* (ours)} & certified TP-Space grid, batched & $O(N\,\bar L)$, indep.\ of $V$, $B$ & certified; $\varepsilon$-suboptimal w.r.t.\ abstraction \\
\bottomrule
\end{tabular}
\end{table*}

\section{Background: PTGs and TP-Space}\label{sec:background}

A \emph{Parameterized Trajectory Generator} (PTG) is a family of trajectories
$\gamma_k : [0, D] \to \SEtwo$, $k\in\{1,\dots,K\}$, that start at the origin
of the vehicle frame and are feasible by construction for the vehicle's
kinematics and velocity bounds. The argument $d$ is a traveled pseudo-distance
that upper-bounds the Euclidean arc length of the reference point, so the
position part of $\gamma_k(d)$ has norm at most $d \le D$. The maximum
pseudo-distance $D$ is a design parameter that sets the planner's short-term
horizon; its exact value is not critical for TPS-A*, as long as it exceeds
the longest search edge (Sec.~\ref{sec:algorithm}). A PTG also provides
the associated body twist and the \emph{TP-Space transform}
$\tau : \R^2 \to \mathcal{T}$, $\tau(x,y) = (k,d)$, which finds the trajectory
and distance at which the reference point reaches $(x,y)$, whatever the
heading: along a trajectory the heading is determined by $d$. The
\emph{Trajectory Parameter Space} $\mathcal{T}$ is the space of $(k,d)$ pairs,
in which the vehicle is a free-flying point~\cite{blanco2006tpspace}.

We use the circular-arc family (\emph{C-PTG}): each $k$ has a constant
curvature bounded by the minimum turning radius, and a sign parameter selects
forward or reverse motion, recovering the Reeds-Shepp-like maneuvers of a
car-like vehicle. Trajectories are generated once, at initialization, by
integrating the steering law with a fixed integration time step (1\,ms of
simulated motion, not computation time) and stored as $M_k{+}1$
samples $d_0{=}0 < d_1 < \dots < d_{M_k}$ with piecewise-constant twist between
consecutive samples, so run-time successor generation never integrates an ODE.
The same holds for PTG families without closed form (e.g., velocity-ramped
holonomic blends).

\section{Certified Collision Checking in TP-Space}\label{sec:collision}

Let $\mathcal{B}\subset\R^2$ be the vehicle footprint, a simple polygon (convex
or not) with $V$ vertices and circumscribed radius $R_\mathcal{B}$ about the
reference point, and let $\mathcal{B}(\Q)$ denote it placed at pose $\Q$. For a
clearance $m \ge 0$, we say that a point $o$ is \emph{touched} at pose $\Q$ if
$\operatorname{dist}(o, \mathcal{B}(\Q)) \le m$, where
$\operatorname{dist}(A, B) = \min_{a\in A,\, b\in B} \lVert a - b\rVert$ is the
Euclidean distance between two sets and the footprint is taken as a closed
region, so the distance is zero for a point inside it.

\subsection{Certified construction}
For each PTG we precompute, offline and once per footprint, a grid $G$ of
cells $c$, each covering the axis-aligned square $S_c\subset\R^2$ of side
$\Delta_c$, that cover the disk of radius
$D + R_\mathcal{B} + m$ in the vehicle frame. Between consecutive samples $n$
and $n{+}1$ of trajectory $k$, the displacement of any footprint point is
bounded by
\begin{equation}
r_{k,n} = \ell_{k,n} + R_\mathcal{B}\,\lvert \phi_{k,n+1} - \phi_{k,n}\rvert,
\label{eq:sweep}
\end{equation}
where $\ell_{k,n}$ is the arc length of the reference point and $\phi$ the
heading ($r_{k,M_k}=0$). The grid stores, for every cell $c$ and trajectory $k$,
\begin{equation}
G[c][k] = \min\bigl\{ d_n :\ \operatorname{dist}\bigl(S_c,\,
\mathcal{B}(\gamma_k(d_n))\bigr) \le r_{k,n} + m \bigr\},
\label{eq:grid}
\end{equation}
with $G[c][k]=D$ if the set is empty (Algorithm~\ref{alg:colgrid}). The
square-to-polygon distance is computed exactly: it is zero if the cell center
lies inside the polygon or a polygon edge crosses the square, and otherwise it
is the minimum over polygon edges of the vertex-to-edge distances between the
two convex sets. In practice each cell keeps a short sparse list of $(k, G[c][k])$
pairs, only for the trajectories that reach it.

\subsection{Run-time query}
At run time, the free distance of trajectory $k$ for a local obstacle cloud
$\{p_j\}_{j=1}^N$, expressed in the vehicle frame of the node being expanded, is
\begin{equation}
\free(k) = \min_j\, G\bigl[\texttt{cell}(p_j)\bigr][k],
\label{eq:dfree-cloud}
\end{equation}
which a single pass over the cloud computes for all $k$ simultaneously
(Algorithm~\ref{alg:colgrid}, bottom). Obstacles farther than
$D + R_\mathcal{B} + m$ from the node cannot be touched by any trajectory and
are clipped away before the pass.

\begin{algorithm}[t]
\caption{Certified TP-Space collision grid}
\label{alg:colgrid}
\small
\begin{algorithmic}[1]
\Statex \textbf{Offline}, once per (footprint, PTG, clearance):
\Require polygon $\mathcal{B}$, clearance $m$, samples $\{\gamma_k(d_n)\}$
\State $G[c][k] \gets D$ for all cells $c$ within radius $D{+}R_\mathcal{B}{+}m$
\ForAll{$k = 1,\dots,K$ and samples $n = 0,\dots,M_k$}
   \State $P \gets \mathcal{B}(\gamma_k(d_n))$;\ $\rho \gets r_{k,n} + m$ \Comment{Eq.~\eqref{eq:sweep}}
   \ForAll{cells $c$ intersecting the bounding box of $P$ grown by $\rho$}
      \If{$\operatorname{dist}(S_c, P) \le \rho$}
         \State $G[c][k] \gets \min\big(G[c][k],\, d_n\big)$
      \EndIf
   \EndFor
\EndFor
\Statex \textbf{Run time}, once per expanded node:
\Require obstacle points $\{p_j\}$ in the node's vehicle frame
\State $\free(k) \gets D$ for all $k$
\ForAll{obstacle points $p_j$}
   \ForAll{$(k, d) \in G[\texttt{cell}(p_j)]$} \Comment{sparse list}
      \State $\free(k) \gets \min\big(\free(k),\, d\big)$
   \EndFor
\EndFor
\State \Return $\free(\cdot)$ \Comment{all $K$ trajectories at once}
\end{algorithmic}
\end{algorithm}

\subsection{Soundness and conservativeness}

\begin{lemma}\label{lem:disp}
For $d \in [d_n, d_{n+1}]$, every point of $\mathcal{B}(\gamma_k(d))$ lies within
$r_{k,n}$ of the corresponding body point of $\mathcal{B}(\gamma_k(d_n))$.
\end{lemma}
\begin{IEEEproof}
A body point $b\in\mathcal{B}$ is at $\mathbf{t}(d) + \mathbf{R}(\phi(d))\,b$.
Its displacement from $d_n$ is at most
$\lVert \mathbf{t}(d)-\mathbf{t}(d_n)\rVert + \lVert b\rVert\,
\lvert\phi(d)-\phi(d_n)\rvert \le \ell_{k,n} + R_\mathcal{B}\lvert
\phi_{k,n+1}-\phi_{k,n}\rvert$, since the twist is constant between samples,
so the heading changes monotonically.
\end{IEEEproof}

\begin{theorem}[Soundness]\label{thm:sound}
Let $o$ be any point and $d^\star$ the smallest distance at which $o$ is touched
along $\gamma_k$. Then $G[\texttt{cell}(o)][k] \le d^\star$. Consequently, an
edge along trajectory $k$ of length $d_e < \free(k)$ keeps every obstacle point
strictly farther than $m$ from the footprint during the whole continuous motion.
\end{theorem}
\begin{IEEEproof}
If $o$ lies outside the grid, then $\lVert o\rVert > D + R_\mathcal{B} + m$, which
exceeds the distance from the origin to any point of the grown footprint along
any trajectory, since the reference point never goes farther than $D$; so $o$
is never touched. Otherwise, let $n$ be such that $d_n \le d^\star < d_{n+1}$
(or $n{=}M_k$). Let $z\in\mathcal{B}(\gamma_k(d^\star))$ with
$\lVert o - z\rVert\le m$, and $z'$ the same body point at $d_n$. By
Lemma~\ref{lem:disp}, $\lVert z-z'\rVert \le r_{k,n}$, hence
$\operatorname{dist}(S_c,\mathcal{B}(\gamma_k(d_n))) \le
\operatorname{dist}(o,\mathcal{B}(\gamma_k(d_n))) \le m + r_{k,n}$ for
$c=\texttt{cell}(o)$, and \eqref{eq:grid} gives $G[c][k] \le d_n \le d^\star$.
The consequence follows from \eqref{eq:dfree-cloud}.
\end{IEEEproof}

\begin{corollary}[Region obstacles]\label{cor:regions}
Let the obstacles be closed sets whose boundaries are sampled by points $\{p_j\}$
such that every boundary point is within $\rho$ of a sample, and let the grid be
built with $m \ge \rho$. If an edge starts collision-free and $d_e < \free(k)$,
the footprint does not intersect any obstacle along the edge.
\end{corollary}
\begin{IEEEproof}
Otherwise let $d^\star$ be the first contact. Since the footprint was disjoint
from the obstacles before $d^\star$ and moves continuously, at $d^\star$ it
meets them only on their boundary, at a point within $\rho \le m$ of some sample
$p_j$; so $p_j$ is touched at $d^\star < \free(k)$, contradicting
Theorem~\ref{thm:sound}.
\end{IEEEproof}

The price of soundness is a bounded conservativeness: by \eqref{eq:grid}, a
point $o$ in cell $c$ can only block trajectory $k$ at $d_n$ if it lies within
$m + r_{k,n} + \sqrt{2}\,\Delta_c$ of the footprint at that sample. With the
1\,ms integration step, $r_{k,n}$ is below $3$\,mm in all our settings, so the
margin is dominated by the cell diagonal and shrinks with $\Delta_c$ at the cost
of a longer offline build. The construction in the reference implementation used by earlier TP-Space
work~\cite{blanco2008extending,blanco2015tpsrrt} marked a cell when one of its
corners fell inside the footprint at a sample, and covered only the radius-$D$
disk; neither choice is conservative. A brute-force test (Sec.~\ref{sec:exp-micro}) confirms
violations for the corner rule and none for \eqref{eq:grid}.

\subsection{Per-node cost}

\begin{proposition}[Per-node collision cost]\label{prop:cost}
Computing $\free(k)$ for all $K$ trajectories of a PTG at one node costs
$O(N\,\bar L)$ operations, where $\bar L \le K$ is the mean length of the
per-cell lists, independently of the vertex count $V$ and of the number $B$ of
successors subsequently taken from the family.
\end{proposition}
\begin{IEEEproof}
Each of the $N$ points triggers one cell lookup and one pass over its list;
neither depends on $V$, which only enters the offline build. Selecting a
successor $(k,d)$ is then an $O(1)$ comparison $d < \free(k)$.
\end{IEEEproof}

A pose-by-pose checker instead costs $O(B\,n_p\,C(V,N))$ per node, where $n_p$
poses are checked per successor and $C(V,N)$ is the cost of one
footprint-versus-obstacles test, which grows with $V$ (or, for a rasterized
check, with the footprint perimeter). Precomputed swept cells remove $V$ but
still pay per successor, $O(B\,S)$: the difference is the direction of the
lookup, from obstacles to motions rather than from motions to cells, which
makes the cost proportional to the local obstacles instead of to the swept
area of every successor. The footprint enters $\bar L$ only
weakly: a larger swept footprint reaches more cells, so lists become slightly
longer; we measure this effect in Sec.~\ref{sec:exp-micro}. The model costs an
offline build that grows linearly with $V$ and with the number of samples, and
a grid whose size is set by $\Delta_c$, $D$, and $K$.

\section{The TPS-A* Algorithm}\label{sec:algorithm}

\subsection{Search space and nodes}
TPS-A* searches $\SEtwo = \R^2 \times S^1$, discretized into cells of size
$(\Delta_{xy}, \Delta_\theta)$ within a bounded world box and stored sparsely.
As in Hybrid-A*~\cite{dolgov2010path}, each cell keeps one \emph{representative}
continuous kinematic state $\X = (\Q,\mathbf{v})$: the end state of the cheapest
edge that has reached it so far. The representative of a cell may be replaced
while the cell is open, and is frozen once the cell is expanded, since expanded
cells are never reopened. Keeping continuous states avoids accumulating
lattice snapping error, so the returned path is an exact concatenation of PTG
segments.

\subsection{Successor generation}
For each PTG $\mathcal{P}_i$, the node's local obstacles are passed once through
the collision grid of Sec.~\ref{sec:collision} to obtain $\free(\cdot)$ for all $k$
(Algorithm~\ref{alg:tps-astar}, lines \ref{ln:cand-begin}--\ref{ln:cand-end}).
A subset of selectors $k$, plus the selector pointing at the goal, is then
combined with a set of time horizons, and every sample $(k,d)$ with $d <
\free(k)$ yields a candidate successor $\X_n \oplus \gamma(k,d)$. If several
candidates reach the same cell, only the shortest one is kept.

\subsection{Cost and heuristic}
The edge cost is the execution time $T(k,d)$ of the PTG segment, optionally
augmented by non-negative cost terms (e.g.\ a clearance penalty). For the
fixed-speed C-PTGs used here it equals arc length over the nominal speed. For a
position goal the heuristic is $h(\X) = \lVert \mathbf{p} - \mathbf{p}_g\rVert /
v_\text{max}$: any motion with speed at most $v_\text{max}$ needs at least that
long to cover the displacement, so $h$ is consistent~\cite{hart1968formal}.
For a full-pose goal and a set of forward and reverse C-PTGs, every edge is a
curve of curvature at most $1/R$, with $R = \min_i v_{\text{max},i}/\omega_{\text{max},i}$ over the PTGs
$\mathcal{P}_i$ of the set,
so the length $L_\text{RS}$ of the shortest Reeds-Shepp
path~\cite{reeds1990optimal} to the goal pose is a lower bound on the length of
any feasible path, and $h(\X) = L_\text{RS}(\Q, \Q_g)/v_\text{max}$ is
admissible; it is also consistent, because $L_\text{RS}$ satisfies the triangle
inequality and never exceeds the length of an edge. Unlike a distance-based
heuristic, it accounts for the reversals a goal heading may require, which is
what Hybrid-A* uses it for~\cite{dolgov2010path}. The open set
is ordered by $f = g + \varepsilon h$, where $g(\X)$ is the cost of the
cheapest path found so far from the start to $\X$ (weighted A*~\cite{pohl1970heuristic}); we
use $\varepsilon{=}1.5$ in all benchmarks.

\subsection{Deferred analytic goal expansion}\label{sec:goal-expansion}
Candidates that land in the goal cell are detected by the successor routine at
no extra cost, since the goal-pointing selector is always sampled. Terminating
on the \emph{first} such edge, as in the analytic expansion of
Hybrid-A*~\cite{dolgov2010path}, is greedy: with a full-pose goal and a tight
turning radius, a node whose heading is one yaw cell off may only re-enter the
goal cell through a near-full-circle arc. TPS-A* instead keeps the cheapest
goal-landing edge as a candidate (Algorithm~\ref{alg:tps-astar},
line~\ref{ln:ae-cand}) and commits to it once its cost does not exceed the
minimum $f$ in the open set (line~\ref{ln:ae-commit}), which preserves the
suboptimality bound (Proposition~\ref{prop:bound}) while retaining most of the
early-termination benefit. Stale open-set entries are discarded lazily.

\textbf{Reeds-Shepp shot.} With a full-pose goal the sampled successors seldom
land in the goal cell with the right heading, so the search alone needs many
expansions near the goal. From every expanded node whose Reeds-Shepp distance
to the goal pose is at most a threshold $L_\text{shot}$, TPS-A* therefore computes the
shortest Reeds-Shepp path~\cite{reeds1990optimal} to the goal pose at the
tightest turning radius of its C-PTGs (Algorithm~\ref{alg:tps-astar},
line~\ref{ln:rs-shot}). Its (at most five) segments are tightest arcs and
straight lines, forward or reverse, which are exactly the extreme and middle
trajectories of the forward and reverse C-PTGs; the shot is thus a chain of
ordinary PTG edges, and each is accepted only if $d < \free(k)$ at its start
pose (segments longer than a trajectory are split). A shot that passes becomes
a goal candidate under the same commit rule, and it ends at the goal pose
itself, not only in the goal cell. This is the analytic expansion of
Hybrid-A*~\cite{dolgov2010path}, but executed and certified with the PTGs.
For a position goal, the shot follows the shortest Reeds-Shepp path to the goal
point with a free final heading. Near obstacles, the footprint may fit at the
goal only with some headings, so a search about to end in the goal cell first
tries, from that node and its ancestors within $L_\text{shot}$, the shortest paths
to every lattice heading at which the footprint, grown by $m$, fits at the goal;
if it fits at none, no shot is tried.

\begin{algorithm}[t]
\caption{TPS-A* path planning}
\label{alg:tps-astar}
\small
\begin{algorithmic}[1]
\Require start $\X_s$, goal $\X_g$, obstacles $\obs$, PTG set $\{\mathcal{P}_i\}$,
heuristic weight $\varepsilon \ge 1$
\Ensure feasible path $\pi$ from $\X_s$ to $\X_g$, or $\varnothing$
\State $g(\cdot)\gets\infty$;\ $g(\X_s)\gets 0$;\ $f(\X_s)\gets \varepsilon\,h(\X_s)$;\ $\textsc{Open} \gets \{\X_s\}$
\State $(\X_{a},\,g_{a}) \gets (\varnothing,\,\infty)$ \Comment{best goal-landing candidate}
\While{$\textsc{Open}\neq\varnothing$}
  \State $\X_n \gets \argmin_{\X\in\textsc{Open}} f(\X)$
  \If{$g_{a} \le f(\X_n)$} \Return $\textsc{Backtrack}(\X_{a})$ \Comment{deferred commit} \label{ln:ae-commit}
  \EndIf
  \If{$\textsc{cell}(\X_n) = \textsc{cell}(\X_g)$} \Return $\textsc{Backtrack}(\X_n)$
  \EndIf
  \State move $\X_n$ from \textsc{Open} to \textsc{Closed}
  \ForAll{$\mathcal{P}_i \in \{\mathcal{P}_i\}$} \label{ln:cand-begin}
     \State $\free(\cdot) \gets$ free-distance query of Alg.~\ref{alg:colgrid} for $\obs$ in the frame of $\Q_n$ \Comment{all $k$}
     \ForAll{sampled $(k,d)$, $k$ incl.\ goal-pointing index}
        \If{$d < \free(k)$ \textbf{and} $\textsc{cell}(\X_n \oplus \gamma(k,d)) \notin \textsc{Closed}$}
           \State $\X' \gets \X_n \oplus \gamma(k,d)$;\ \ $c \gets g(\X_n)+T(k,d)$
           \If{$c < g(\textsc{cell}(\X'))$} \Comment{replaces the representative}
              \State $g(\X')\gets c$;\ $f(\X')\gets c + \varepsilon\,h(\X')$
              \State $\text{parent}(\X')\gets \X_n$;\ insert $\X'$ in \textsc{Open}
              \If{$\textsc{cell}(\X') = \textsc{cell}(\X_g)$ \textbf{and} $c < g_{a}$}
                 \State $(\X_{a},\,g_{a}) \gets (\X',\,c)$ \Comment{goal candidate} \label{ln:ae-cand}
              \EndIf
           \EndIf
        \EndIf
     \EndFor \label{ln:cand-end}
  \EndFor
  \If{Reeds-Shepp shot from $\X_n$ is certified, of cost $c < g_a$} \label{ln:rs-shot}
     \State $(\X_{a},\,g_{a}) \gets$ (end of the shot, $c$) \Comment{Sec.~\ref{sec:goal-expansion}}
  \EndIf
\EndWhile
\State \Return $\varnothing$
\end{algorithmic}
\end{algorithm}

\section{Theoretical Properties}\label{sec:theory}

We now state what TPS-A* guarantees. Let $\mathcal{G}$ be the lattice
abstraction explored by a run: its vertices are the cells of the world box and
the out-edges of a cell are the certified successors generated from its
representative state (for expanded cells, the frozen one), including its
Reeds-Shepp shot, if any.

\begin{theorem}[Certified paths]\label{thm:paths}
If the start state is collision-free, every path returned by TPS-A* is a
concatenation of PTG segments, hence satisfies the vehicle's kinematic and
velocity constraints, and keeps the footprint, grown by the clearance $m$,
disjoint from every obstacle point during the entire continuous motion; for
region obstacles sampled as in Corollary~\ref{cor:regions}, the footprint never
intersects them.
\end{theorem}
\begin{IEEEproof}
Every accepted edge satisfies $d < \free(k)$ at the pose where it starts, so
Theorem~\ref{thm:sound} (resp.\ Corollary~\ref{cor:regions}) applies to it, and
its end state is collision-free, which is the premise for the next edge. The
claim follows by induction along the path. Feasibility holds because each edge
is a PTG trajectory.
\end{IEEEproof}

TPS-A* also terminates, after at most as many expansions as there are cells in
the world box, since expanded cells are never reopened.

\begin{proposition}[Bounded suboptimality]\label{prop:bound}
For a position goal with the Euclidean heuristic, or a full-pose goal with the
Reeds-Shepp heuristic and C-PTGs (Sec.~\ref{sec:algorithm}), and non-negative
edge costs, the cost $C$ of the returned path satisfies
$C \le \varepsilon\,(C^\star_{\mathcal{G}} + \delta)$, where
$C^\star_{\mathcal{G}}$ is the cost of the cheapest start-to-goal-cell path in
$\mathcal{G}$ and $\delta = \max_{\X \in \text{goal cell}} h(\X)$ accounts for
reaching the goal \emph{cell} rather than the goal itself: $\delta =
\sqrt{2}\,\Delta_{xy}/v_\text{max}$ for a position goal.
\end{proposition}
\begin{IEEEproof}
Both heuristics are consistent and satisfy the triangle inequality, and the
goal lies in the goal cell, so $h$ overestimates the time to reach that cell by
at most $\delta$. The classical argument for weighted A*
without re-expansions~\cite{likhachev2003ara} then applies on $\mathcal{G}$: out-edges
of expanded cells never change after expansion, and until termination the open
set contains a cell $u$ of an optimal path of $\mathcal{G}$ with
$f(u) \le \varepsilon\,(C^\star_{\mathcal{G}} + \delta)$, so
$\min_\textsc{Open} f \le \varepsilon\,(C^\star_{\mathcal{G}} + \delta)$ at
every iteration. If the search stops by popping the goal cell,
$C \le f \le \min_\textsc{Open} f$; if it commits the deferred candidate
(line~\ref{ln:ae-commit}), $C = g_a \le \min_\textsc{Open} f$. Both are bounded
as claimed.
\end{IEEEproof}

\textbf{What is not guaranteed.} As in Hybrid-A*~\cite{dolgov2010path}, keeping
one representative per cell means that $\mathcal{G}$ depends on the order of
expansions: a feasible motion through a cell may be lost if the cell's
representative, which differs from that motion's state by less than one cell,
cannot continue it. TPS-A* is therefore complete and $\varepsilon$-suboptimal
\emph{relative to its abstraction}, not resolution complete in the strict
sense; state lattices whose primitives end on lattice states avoid this at the
cost of a fixed, offline primitive set. Proposition~\ref{prop:bound} does not
cover a full-pose goal with a non-admissible heuristic, and for pose goals
$\delta$ is not small: a lateral offset within one cell can only be corrected
by an S-shaped maneuver, so for the car of Sec.~\ref{sec:exp-parking}
($R{=}3$\,m, $\Delta_{xy}{=}0.10$\,m) $\delta \approx 1.5$\,m of travel.
Sec.~\ref{sec:exp-barn} reports the realized suboptimality and the rate of
unsolved queries.

\section{Experimental Validation}\label{sec:experiments}

We evaluate, in order: (i) the per-node cost of the collision check against
footprint complexity and branching factor, and the soundness of its
construction; (ii) success, latency, and path quality on a neutral benchmark
against strong baselines, including the realized suboptimality of
$\varepsilon{=}1.5$; (iii) the any-shape benefit on narrow doorways; (iv)
automated parking with a passenger car; (v) docking with a non-convex
footprint; and (vi) a real wheeled robot navigating a university campus.

\subsection{Setup}\label{sec:exp-setup}
All experiments but the real-robot one (Sec.~\ref{sec:exp-robot}) are
\emph{planning only}: perception, localization, and control are excluded to
isolate the planner and its collision model. The baselines span
the two families of car-like global planners of the Bench-MR
benchmark~\cite{heiden2021benchmr}: search-based Nav2 Smac
Hybrid-A*~\cite{macenski2023survey} and SBPL ARA*~\cite{likhachev2003ara} on an
$(x,y,\theta)$ lattice, and sampling-based OMPL
RRT*/BIT*~\cite{sucan2012ompl} with a Reeds-Shepp connector and
FCL~\cite{pan2012fcl} footprint collision. Each runs standalone on the
\emph{same} occupancy data, with the \emph{same} polygon footprint and minimum
turning radius as TPS-A* and identical queries, treating each occupied cell at
its full square extent.

\textbf{Certified obstacle model for TPS-A*.} TPS-A* consumes points: following
Corollary~\ref{cor:regions}, occupied regions are represented by points every
$0.05$\,m along the occupied/free boundary ($\rho = 0.025$\,m), with grids
built with $m = \rho$, so every TPS-A* path is certified against the full
occupied squares; the local obstacle window is the reach $D + R_\mathcal{B} +
m$.

\textbf{Independent verification.} Every returned path of every planner is
checked against the raw occupancy grid by an exact separating-axis test between
the footprint polygon and each occupied cell square, sharing no code with any
planner: at every PTG sample for TPS-A* (millimeter spacing), every primitive
sample for SBPL, and every $1$\,cm for OMPL. Nav2's returned poses, about
$0.23$\,m apart, are checked as returned, so its count of colliding paths is a
lower bound.

\textbf{Protocol.} We report success, planning wall-time (the planner's own
\texttt{plan()} call), and path length, with paired Wilcoxon signed-rank tests
on the queries solved by both planners. Each planner runs once per query,
single-threaded and serially, as an optimized build on an AMD Ryzen~7 2700X
under Linux. OMPL is anytime, so it is compared on success and path quality
only. TPS-A* uses a forward+reverse C-PTG pair with $41$ trajectories each,
$\Delta_c = 0.05$\,m, $\Delta_\theta = 7.5^\circ$, $\varepsilon = 1.5$, the
Reeds-Shepp shot ($L_\text{shot}{=}5$\,m) and, for pose goals, the Reeds-Shepp
heuristic, and on
BARN $\Delta_{xy}=0.15$\,m, the map resolution; the other experiments state
their changes.

\subsection{Collision checking: cost and soundness}\label{sec:exp-micro}
\textbf{Setup.} A single forward C-PTG ($K{=}121$, $D{=}5$\,m,
$\Delta_c{=}0.05$\,m) is paired with regular-polygon footprints of $4$ to $64$
vertices. Two baselines answer the same query, the collision-free distance
along trajectory $k$, pose by pose at a $0.10$\,m arc step with early exit:
FCL~\cite{pan2012fcl} (convex prism, broadphase AABB tree) and Nav2's
\texttt{FootprintCollisionChecker} on a rasterized cost map, the check inside
\texttt{SmacPlannerHybrid}. Clouds of $200$ to $20{,}000$ points form $10$ fixed
clusters, so that more points densify each obstacle rather than fill the free
lanes; $200$ points match the median number of occupied cells within $5$\,m
across $20$ BARN worlds, one point each (the planner's certified boundary
sampling roughly triples it, Sec.~\ref{sec:exp-barn}).

\begin{figure}[t]
\centering
\includegraphics[width=\columnwidth]{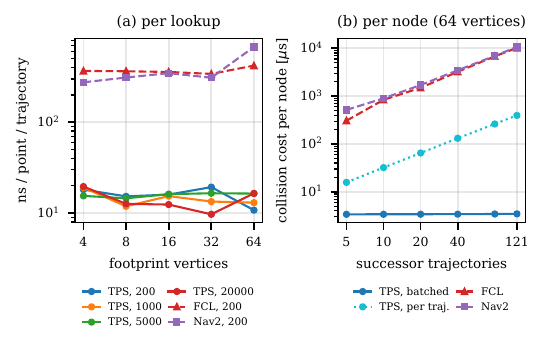}
\caption{Collision-checking microbenchmark (forward C-PTG, $121$ trajectories,
$\Delta_c{=}0.05$\,m). (a) Cost per obstacle point and trajectory, one
trajectory at a time, vs.\ footprint vertices; legend: checker, number of
obstacle points. (b) Collision cost of one node expansion vs.\ the number of
successor trajectories ($200$ points).}
\label{fig:micro}
\end{figure}

\textbf{Cost per node.} Fig.~\ref{fig:micro}(b) shows the quantity that matters
for search, the collision cost of one node expansion as a function of the number
$B$ of successor trajectories, with $200$ obstacle points. The node-batched query
costs $3.1$\,$\mu$s with the $4$-vertex footprint and $3.4$--$3.5$\,$\mu$s
with the $64$-vertex one, flat in $B$ and in $V$, as
Proposition~\ref{prop:cost} predicts. Querying the same grid one trajectory at a
time grows linearly with $B$ ($16$ to $397$\,$\mu$s), and so do the pose-by-pose
baselines: from $0.29$--$0.31$ to $8.8$--$10.2$\,ms (FCL) and $0.23$--$0.52$ to
$6.5$--$10.7$\,ms (Nav2) as $B$ grows from $5$ to $121$. At $B{=}20$, the value
used by our planner per PTG, the batched query is about $440\times$ and
$500\times$ cheaper than FCL and Nav2 with the $64$-vertex footprint.

\textbf{Cost per lookup and footprint complexity.} Per obstacle point and
trajectory, queried one at a time (Fig.~\ref{fig:micro}(a)), TPS costs
$11$--$19$\,ns across $4$--$64$ vertices, versus $341$--$421$\,ns for FCL and
$275$--$675$\,ns for Nav2 at $200$ points: $15$--$62\times$ cheaper even
without batching. TPS visits every obstacle point, however, whereas Nav2's
rasterized lookup does not depend on the point count, so at $5$k--$20$k points
in the $5$\,m window, far beyond the BARN-calibrated range, Nav2's per-point
cost reaches and then drops below TPS's. The one-time cost is a
$5$--$8$\,s offline build with no systematic growth in $V$ and a grid of about
$0.95$\,MB.

\textbf{Soundness.} A brute-force test in the library sweeps an L-shaped
footprint along every trajectory of a C-PTG at sub-sample resolution against
random points: the certified grid never exceeds the true first contact, with or
without clearance, whereas the previous corner-test construction fails ($3$
violations in $86$ contact pairs, two from obstacles beyond the radius-$D$
grid), consistent with the $5$ colliding BARN paths of Sec.~\ref{sec:exp-barn}.

\subsection{BARN: a level-field comparison}\label{sec:exp-barn}
The BARN dataset~\cite{perille2020barn,xiao2022barn} provides $300$ cluttered $30{\times}30$
worlds for the Clearpath Jackal. All planners use its true rectangular footprint
($0.508 \times 0.430$\,m) and turning radius $0.382$\,m (Reeds-Shepp expansions
for Nav2 and OMPL, forward and reverse C-PTGs for TPS-A*). Following BARN's
goal-region scoring, the nominal goal is snapped once per world to the nearest
position within $0.5$\,m with circumscribed-radius clearance (unchanged in
$221$ worlds). BARN goals are positions; we report (i) a \emph{shared goal
pose} with the heading of travel ($90^\circ$), since SBPL, OMPL's $\SEtwo$
space, and Nav2 up to Jazzy need one, and (ii) BARN's \emph{position goal} for
the planners that support it (Nav2 from Kilted on, OMPL with a goal region; not
SBPL). Terminal acceptance: TPS-A*, its goal cell ($0.15$\,m, $7.5^\circ$) or
the goal itself after a Reeds-Shepp shot; Nav2 and OMPL, $0.25$\,m; SBPL,
its goal lattice state. Nav2 runs in its documented configuration, with the
inflation layer (radius $0.55$\,m, cost scaling $3$) its footprint check relies
on, in the same session as TPS-A*. SBPL runs ARA* in first-solution mode
(initial $\varepsilon{=}3$) over forward and reverse unicycle primitives for
the matched radius, checking the exact polygon against un-inflated cells;
OMPL uses FCL (an oriented box, exact here) and a $2$\,s budget.

\begin{figure*}[t]
\centering
\includegraphics[width=\textwidth]{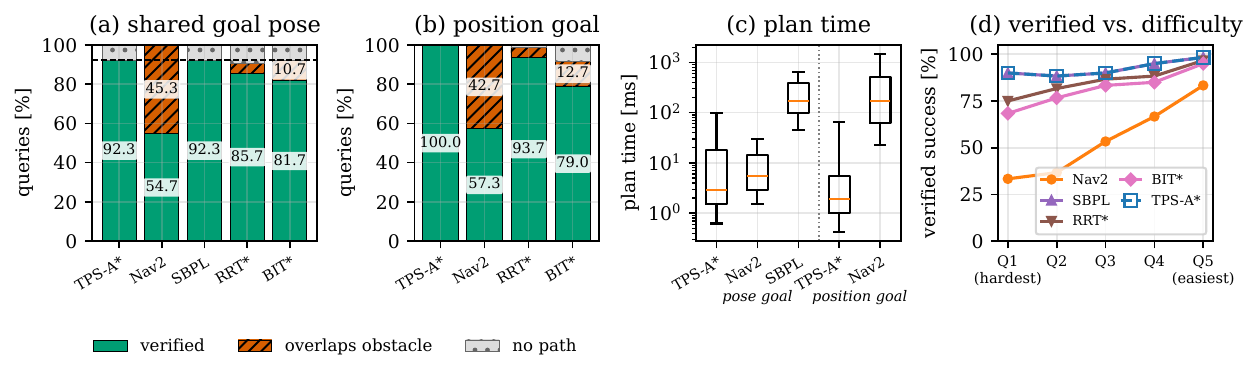}
\caption{BARN, $300$ worlds. (a,b) Outcome of every query with the shared goal
pose and with the position goal: verified path (solid, value shown), path
overlapping an obstacle (hatched), no path (dotted); dashed line: the $277$
collision-free goal poses. (c) Plan time of the deterministic planners (solved
queries). (d) Verified success with the shared goal pose by BARN difficulty
quintile (Q1: tightest); TPS-A* coincides with SBPL. RRT*, BIT*: OMPL.}
\label{fig:barn}
\end{figure*}

\textbf{Shared goal pose.} Table~\ref{tab:barn} and Fig.~\ref{fig:barn}
summarize the result. The exact test shows that in $23$ worlds the footprint at
the goal pose already overlaps an occupied cell, so at most $277$ queries
($92.3\%$) admit a collision-free solution. TPS-A* solves all $277$ and
returns no path for the $23$ infeasible ones; all its paths pass the exact
independent verification ($2.09$\,M poses, no overlap with any occupied cell),
as Theorem~\ref{thm:paths} predicts. SBPL behaves identically, and OMPL returns
a path for $271$ (RRT*) and $277$ (BIT*) of the feasible queries. Nav2 returns
a path for all $300$, including the $23$ infeasible ones, but many of its paths
collide (see below): only $54.7\%$ of the queries receive a verified path, and
$33.3$--$36.7\%$ in the two hardest difficulty quintiles
(Fig.~\ref{fig:barn}(d), BARN distance-to-closest-obstacle metric), where
TPS-A* and SBPL keep $88.3$--$90\%$, the feasible share. On the $277$
feasible queries TPS-A* takes a median of $2.9$\,ms (mean $24$\,ms, maximum
$0.66$\,s), on par with Nav2 ($5.1$\,ms, mean $10$\,ms; paired Wilcoxon
$p \approx 0.3$) and $58\times$ faster than SBPL in median ($169$\,ms,
$p \approx 10^{-47}$), the other planner whose paths all verify. Its paths are
$2\%$ and $3\%$ longer in median than those of Nav2 and SBPL, an effect of
$\varepsilon{=}1.5$ and of the coarse, execution-time-optimal primitive
sampling. The Reeds-Shepp shot (Sec.~\ref{sec:goal-expansion}) is what makes
pose goals cheap: without it, TPS-A* needs $722$ instead of $14$ expansions
in median, takes $68$\,ms, and its paths are $6\%$ longer.

\textbf{Position goal.} With BARN's original goal, TPS-A* solves all $300$
worlds, the $23$ above included, with all paths verified, in $1.9$\,ms median
($12$ expansions; mean $20$\,ms, maximum $0.69$\,s). In $274$ worlds its path
ends at the goal point itself, within $5$\,mm (median $0.7$\,mm, the rounding
to PTG time steps); in the $23$ above, the footprint grown by $m$ fits at no
heading at the goal, and these and $3$ other paths end in the goal cell. Nav2
(Kilted, free goal heading) is $89\times$ slower in median ($170$\,ms, mean
$391$\,ms; $p \approx 10^{-49}$) and verifies $57.3\%$ of its paths; TPS-A*
paths are $2\%$ longer. OMPL RRT* and BIT* return $296$ and $275$ paths within their
$2$\,s budget, $281$ and $237$ of them verified. BARN does not exercise the
any-shape advantage: its worlds are open fields of cylinders that seldom
require a specific heading to pass an opening; the any-shape benefit is shown
in Sec.~\ref{sec:exp-house} and~\ref{sec:exp-dock}.

\textbf{Where the time goes.} With position goals, the library's profiler
splits the planning time as follows: the node-batched collision query $38\%$,
clipping and transforming the local obstacles $15\%$, enumerating, evaluating,
and gathering candidates $19\%$, the Reeds-Shepp shots $12\%$, other open-set
bookkeeping $13\%$, and setup $4\%$. A node costs about $150\,\mu$s, of which
the collision query takes $36\,\mu$s for both PTGs, more than in
Sec.~\ref{sec:exp-micro} because certified boundary sampling puts about $630$
points (median) within the reach of a node, against $200$ there. With pose
goals, the shots take $30\%$ of the time, including their collision checks,
and a node costs about $230\,\mu$s. Either way, only $12$--$14$ nodes are
expanded in median. The time per query is thus set by
the number of expansions and, within a node, by the number of local points
(Proposition~\ref{prop:cost}), not by the footprint.

\textbf{Price of certification.} The margin is set by the grid cell
(Sec.~\ref{sec:collision}): with $\Delta_c{=}0.10$\,m, TPS-A* solves
$288/300$ worlds with position goals, losing $12$ whose narrowest passage the
coarser margin no longer admits, all paths verified. At $\Delta_c{=}0.05$\,m,
the offline build of both PTGs takes about $3$\,s and each grid file is about
$0.13$\,MB. With the previous, non-certified construction and a corner-point
obstacle model, $5$ of the $300$ paths penetrate an occupied cell by up to
$3.9$\,cm. Relative to that corner-point model, boundary sampling roughly
doubles the number of obstacle points,
which cost about $20\%$ of planning time when certification was introduced.

\textbf{Are the baselines' paths collision-free?} Not always
(Table~\ref{tab:barn}, Fig.~\ref{fig:barn}(a,b)). SBPL's paths all pass;
OMPL's overlap occupied cells by at most $2.2$\,cm, since its motions are
validated at discrete states ($0.5\%$ of the state-space extent apart in our
setup). With the shared goal pose, $136$ of Nav2's $300$ paths ($45\%$) overlap
an occupied cell at a returned pose, $127$ by more than $1$\,cm and up to
$15.5$\,cm, including all $23$ infeasible queries ($128$ with the position
goal). The smoother is not the main cause: without it, $98$ paths still
overlap. \texttt{SmacPlannerHybrid} rasterizes the footprint outline on the
$0.15$\,m cost map at the cell-snapped node position, and the inflation layer
rejects a pose only when an obstacle lies within the inscribed radius of its
center; without the inflation layer, Nav2 solves $241$ queries and $155$ of its
paths overlap. TPS-A* needs no margin: its check is sound by construction
(Theorem~\ref{thm:sound}).

\textbf{Clearance and costs.} Nav2 always plans on its inflated costmap; TPS-A*
needs no cost layer for safety, but has an analogous obstacle-proximity cost,
evaluated with its library defaults (``costmap'', Table~\ref{tab:barn}).
Without it, TPS-A* already keeps more clearance than every baseline; with it,
$9.7$\,cm, with all paths verified, $1$--$3\%$ longer, fewer ending at the goal
pose itself ($229$ vs.\ $271$), and $26$--$41\times$ more expansions, since the
time-only heuristic ignores the cost: $91$\,ms with pose goals (Nav2, $5$\,ms;
SBPL, $169$\,ms), $47$\,ms with position goals (Nav2, $170$\,ms). A lighter
setting adds $1$\,cm of clearance for $1.5\times$ the time.

\textbf{Design choices.} Table~\ref{tab:ablation} changes one setting at a time
in a separate, same-session run, with position goals. $\varepsilon$ matters most:
exact A* ($\varepsilon{=}1$) needs $24\times$ more expansions in median and is
an order of magnitude slower. The cost of the $\varepsilon{=}1.5$ path relative
to the $\varepsilon{=}1$ path is $1.026$ in median, $1.124$ at the 95th
percentile, and $1.29$ at most, well within the factor $1.5$ of
Proposition~\ref{prop:bound}; in $9$ worlds it is below one, because the two
runs explore different representatives and hence different abstractions
$\mathcal{G}$ (Sec.~\ref{sec:theory}). Angular resolution
matters little between $5^\circ$ and $7.5^\circ$, while $10^\circ$ loses one
world; dropping the reverse PTG loses one world and doubles the worst-case
time. Without the deferred analytic expansion, which the shots rely on, twice
as many nodes are expanded; the obstacle-aware heuristic has no measurable
effect on BARN, whose goals lie in open fields.

\begin{table}[t]
\centering
\caption{Design choices of TPS-A* on BARN (position goals), one setting
changed at a time in one interleaved run. Reference: $\Delta_\theta{=}7.5^\circ$,
forward+reverse PTGs, $\varepsilon{=}1.5$, analytic expansion and
Reeds-Shepp shots on (they need the reverse PTG and the analytic expansion),
obstacle-aware heuristic off.}
\label{tab:ablation}
\small
\setlength{\tabcolsep}{3pt}
\begin{tabular}{lcccc}
\toprule
Variant & Success & \multicolumn{2}{c}{Time [ms]} & Expansions \\
        &         & median & max & median \\
\midrule
Reference & 100\% & 1.9 & 691 & 12 \\
$\Delta_\theta = 5^\circ$                    & 100\%   & 1.9 & 754 & 12 \\
$\Delta_\theta = 10^\circ$                   & 99.7\%  & 1.9 & 673 & 12 \\
Forward PTG only                             & 99.7\%  & 1.6 & 1491 & 27 \\
No analytic expansion                        & 100\%   & 2.3 & 692 & 25 \\
Obstacle-aware heuristic                     & 100\%   & 1.9 & 690 & 12 \\
$\varepsilon = 1$                            & 100\%   & 31.1 & 1872 & 292 \\
\bottomrule
\end{tabular}
\end{table}

\begin{table}[t]
\centering
\caption{BARN, $300$ worlds, with a goal pose shared by all planners ($23$
of these poses are in collision: $92.3\%$ is the maximum) and with BARN's
position goal. ``Verified'': the path also passes the exact collision check (as
a share of all queries). Times over each planner's solved queries; OMPL runs to
its $2$\,s budget. Clr.: median minimum footprint-to-obstacle distance of the
paths ($0$ if overlapping; Nav2 at its poses only).}
\label{tab:barn}
\small
\setlength{\tabcolsep}{1.6pt}
\begin{tabular}{lcccccc}
\toprule
Planner & Success & Verified & \multicolumn{2}{c}{Time [ms]} & Len. & Clr. \\
        &         &          & median & mean & [m] & [cm] \\
\midrule
\multicolumn{7}{l}{\emph{Shared goal pose}} \\
TPS-A* (ours)            & 92.3\%  & \textbf{92.3\%} & \textbf{2.9} & 24.1 & 7.34 & 5.9 \\
TPS-A*, no shot          & 92.3\%  & 92.3\%  & 68.2 & 142 & 8.42 & 7.6 \\
TPS-A*, costmap          & 92.3\%  & 92.3\%  & 90.6 & 606 & 7.48 & \textbf{9.7} \\
Nav2 Hybrid-A* (Jazzy)   & 100\%   & 54.7\%  & 5.4 & 10.3 & 7.11 & 1.4 \\
Nav2, no smoother        & 100\%   & 67.3\%  & 5.2 & \textbf{10.2} & 7.35 & 3.5 \\
SBPL ARA*                & 92.3\%  & \textbf{92.3\%} & 169 & 256 & \textbf{6.96} & 3.5 \\
OMPL RRT*                & 90.7\%  & 85.7\%  & \multicolumn{2}{c}{($2000$)} & 7.04 & 2.3 \\
OMPL BIT*                & 92.3\%  & 81.7\%  & \multicolumn{2}{c}{($2000$)} & 7.27 & 1.6 \\
\midrule
\multicolumn{7}{l}{\emph{Position goal}} \\
TPS-A* (ours)            & 100\%   & \textbf{100\%} & \textbf{1.9} & \textbf{20.2} & 7.35 & 6.4 \\
TPS-A*, costmap          & 99.7\%  & 99.7\%  & 47.3 & 406 & 7.37 & \textbf{9.3} \\
Nav2 Hybrid-A* (Kilted)  & 100\%   & 57.3\%  & 170 & 391 & 7.05 & 1.7 \\
Nav2, no smoother        & 100\%   & 67.7\%  & 170 & 391 & 7.27 & 3.5 \\
OMPL RRT*                & 98.7\%  & 93.7\%  & \multicolumn{2}{c}{($2000$)} & \textbf{6.84} & 1.6 \\
OMPL BIT*                & 91.7\%  & 79.0\%  & \multicolumn{2}{c}{($2000$)} & 7.37 & 1.5 \\
\bottomrule
\end{tabular}
\end{table}

\subsection{Narrow doorways: the any-shape advantage}\label{sec:exp-house}
To exercise exact-footprint planning we built a benchmark from the
HouseExpo~\cite{li2020houseexpo} dataset of real residential floor plans: $40$
maps, each rasterized to a solid wall point cloud, with a start and goal in two
rooms whose connecting path crosses a narrow doorway (bottleneck $0.70$--$1.10$\,m,
median $0.80$\,m). The robot is an \emph{elongated} differential-drive vehicle,
$0.95 \times 0.55$\,m: its inscribed disk (radius $0.275$\,m) clips door
jambs, its circumscribed disk (diameter $\approx\!1.10$\,m) rejects every door
narrower than that, and the true footprint passes a $0.80$\,m door only when
correctly aligned.

\begin{figure}[t]
\centering
\includegraphics[width=\linewidth]{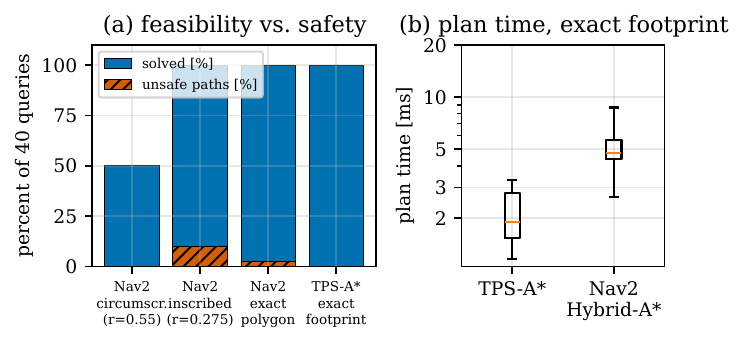}
\caption{HouseExpo doorways ($40$ maps, elongated footprint). (a) A
circumscribed disk rejects half the feasible queries; an inscribed disk yields
$10\%$ unsafe paths; Nav2 with the exact polygon and its inflation layer, one;
TPS-A*, none. (b) Plan time, both planners with the exact polygon.}
\label{fig:house}
\end{figure}

TPS-A* treats the wall pixels ($0.05$\,m squares) as region obstacles
($m{=}0.036$\,m, Corollary~\ref{cor:regions}) and, for these pose goals, uses
the Reeds-Shepp heuristic and shot ($R = 0.25$\,m). With the \emph{same} exact
polygon and its inflation layer, Nav2 also solves all $40$ doorways; TPS-A* is
faster (median $1.9$ vs.\ $4.8$\,ms, $p \approx 3{\times}10^{-11}$), with paths of
indistinguishable length ($p = 0.20$). One Nav2 path overlaps a wall, by $3.6$\,cm ($29$\,cm, another
one, without the inflation layer); all $40$ TPS-A* paths pass the exact
verification ($0.48$\,M poses). The any-shape value is thus not a success-rate
win over a \emph{fairly configured} baseline, but the removal of the lossy
choice an inflated circular model forces (Fig.~\ref{fig:house}(a)): with the
\emph{circumscribed} radius Nav2 solves only $20/40$ queries, rejecting half the
doors the true footprint fits; with the \emph{inscribed} radius it solves all
$40$, but $4$ of those paths collide with the true footprint swept along them.
Without the shot, TPS-A* needs $18.9$\,ms in median.

\subsection{Automated parking}\label{sec:exp-parking}
The $20$ cases of the TPCAP automated-parking benchmark~\cite{li2023tpcap}
cover parallel, perpendicular, and angled parking among polygonal obstacles,
several with slots only $0.2$--$0.4$\,m wider than needed. The car is the
TPCAP vehicle, a $4.69\times1.94$\,m rectangle referenced at the rear axle,
with minimum turning radius $R = 3.0$\,m, so its constant-steering motions are
exactly the arcs of a forward and a reverse C-PTG with
$v_\text{max}/\omega_\text{max}=R$, as they are Reeds-Shepp arcs for Hybrid-A*.
We address the path-planning part of TPCAP (reaching the goal pose
collision-free), leaving the steering-rate and speed profiles it also scores to
a trajectory optimizer, in the benchmark's reference box ($8$\,m around start
and goal). TPS-A* samples the polygon boundaries every $0.05$\,m
($m=0.026$\,m) and uses $\Delta_{xy}=0.10$\,m, successors of $0.25$ to $4$\,m,
and the Reeds-Shepp heuristic and shot ($L_\text{shot}{=}20$\,m, the length of a
maneuver of this car); Nav2 uses the filled polygons on $0.05$\,m cells and
its inflation layer. Both get a $20$\,s budget (success under smaller budgets
is read from the recorded times), without Nav2's default cap of $10^6$
iterations, with which it solves only $12$ cases. Paths are verified exactly
against the original polygons.

\begin{table}[t]
\centering
\caption{TPCAP parking ($20$ cases, $20$\,s budget). Solved within $1$, $5$,
$20$\,s; ``Verif.'': paths overlapping no obstacle; time, length, cusps
(direction switches): medians over the $19$ cases both solve; goal error:
maximum final position/heading error.}
\label{tab:parking}
\footnotesize
\setlength{\tabcolsep}{2pt}
\begin{tabular}{lcccccc}
\toprule
Planner & Solved & Verif. & Time & Len. & Cusps & Goal err. \\
 & $1/5/20$\,s & & [s] & [m] & & [m\,/\,$^\circ$] \\
\midrule
TPS-A* (ours)            & $11/16/19$ & \textbf{19/19} & 0.76 & 19.2 & 3 & 0.11\,/\,4.7 \\
Nav2 Hybrid-A*           & $8/16/19$ & 10/19 & 1.61 & 18.8 & 4 & 0.12\,/\,2.5 \\
Nav2, no smoother        & $8/15/19$ & 19/19 & 1.76 & 18.9 & 4 & 0.12\,/\,2.5 \\
\bottomrule
\end{tabular}
\end{table}

\begin{figure}[t]
\centering
\includegraphics[width=\linewidth]{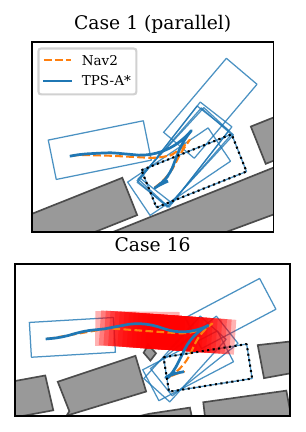}
\caption{TPCAP cases 1 and 16. Blue: TPS-A* path, with the car at the start,
cusps, and end; dotted: goal; dashed: Nav2 path; red: Nav2 poses overlapping an
obstacle (case~16, after its smoother).}
\label{fig:tpcap}
\end{figure}

TPS-A* solves $19$ cases (Table~\ref{tab:parking}, Fig.~\ref{fig:tpcap}), all
verified, $11$ within $1$\,s; the case it misses, also missed by Nav2,
exhausts the budget ($18$ cases without the shot). Nav2 also solves $19$, with
statistically indistinguishable time (median $0.76$ vs.\ $1.61$\,s, $p=0.92$),
length ($p=0.54$), and number of maneuvers ($p=0.17$). Nine of Nav2's default
paths overlap an obstacle, by up to $0.21$\,m, all due to its smoother (none
without it). $16$ TPS-A* paths end with a Reeds-Shepp shot, at the goal pose
itself, the others within $0.11$\,m and $4.7^\circ$; Nav2 lands on the goal
except in three cases where it falls back on its $0.25$\,m tolerance. The
large polygons and the workspace box put about $4{,}400$ points within the
reach of a node, seven times more than on BARN (Proposition~\ref{prop:cost}).

\subsection{Non-convex footprint: fork docking}\label{sec:exp-dock}
\begin{figure}[t]
\centering
\includegraphics[width=\linewidth]{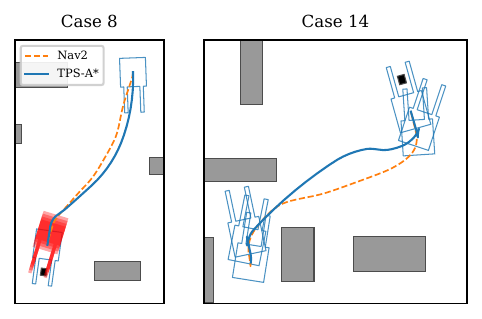}
\caption{Fork docking, non-convex footprint (block in black). Blue: TPS-A*
path, with the footprint at the start, cusps, and end; dashed: Nav2 path; red:
Nav2 poses overlapping an obstacle (case~8, after its smoother).}
\label{fig:docking}
\end{figure}

We generated $20$ warehouse scenes ($12{\times}10$\,m, seven random racks) in
which a fork-style vehicle, a $1.0{\times}0.9$\,m body with two $0.9$\,m tines
$0.44$\,m apart (a non-convex $12$-vertex polygon), must end with its tines
around a $0.25{\times}0.20$\,m block, as when picking up a pallet, from a random
start at least $4$\,m away (forward and reverse C-PTGs, $R{=}0.6$\,m). At every
goal pose the convex hull of the footprint contains the block, so no planner
with a convex footprint model, such as FCL with a convex prism, can solve any
of these queries. TPS-A* uses the exact polygon ($0.05$\,m boundary sampling,
$m{=}0.026$\,m, $\Delta_{xy}{=}0.10$\,m, offline build about $3$\,min) with the
Reeds-Shepp heuristic and shot, and Nav2 the same polygon with its inflation
layer, both with a $20$\,s budget. Both solve all $20$ queries and end with the
block between the tines (Fig.~\ref{fig:docking}); all TPS-A* paths verify,
while $7$ of Nav2's overlap an obstacle, by up to $6$\,cm, all due to its
smoother. Nav2 is faster (median $0.10$ vs.\ $0.26$\,s,
$p \approx 7{\times}10^{-4}$), with $7\%$ shorter paths; $16$ TPS-A* paths end
at the goal pose itself, the others within $0.09$\,m and $5.2^\circ$, and
Nav2's within $0.24$\,m and $6.5^\circ$.

\subsection{Real-robot experiment: Husky on campus}\label{sec:exp-robot}
\begin{figure*}[t]
\centering
\begin{minipage}[b]{0.40\textwidth}
\centering
\includegraphics[height=5.6cm]{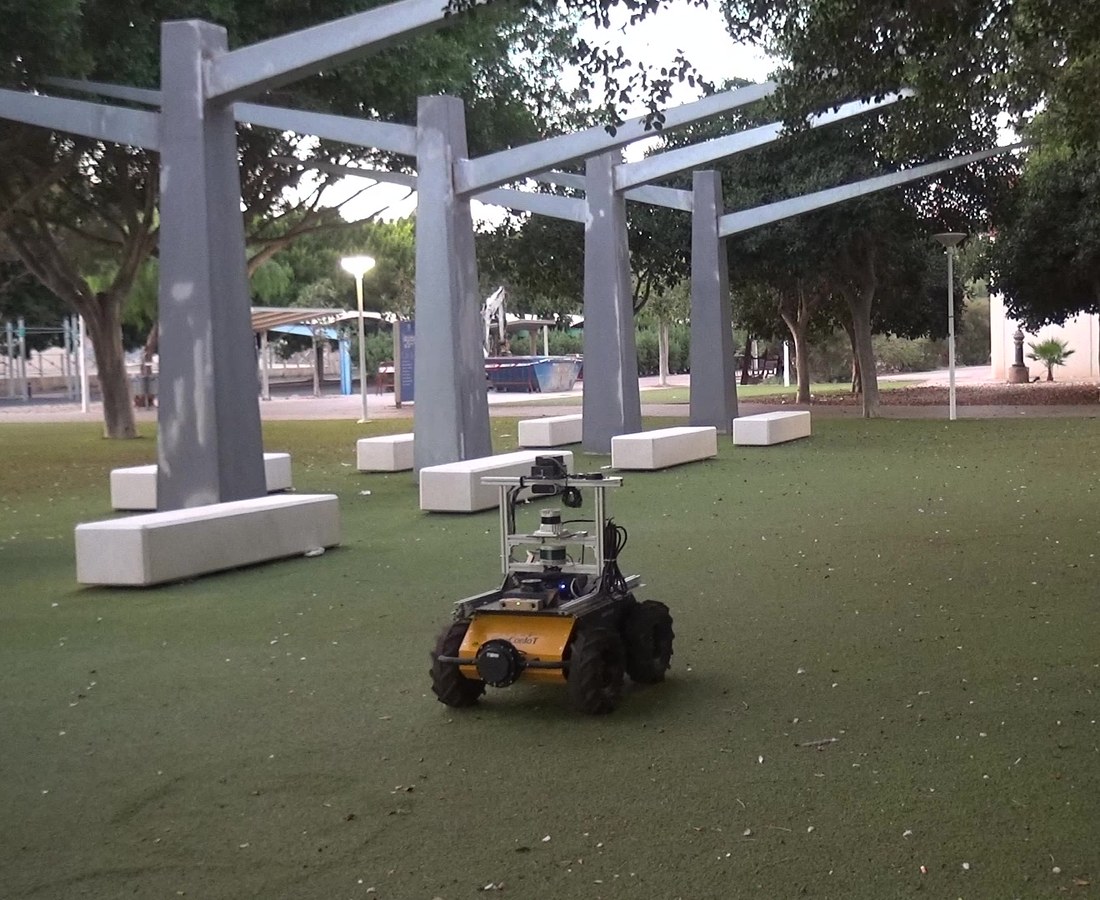}\\[2pt]
\small (a)
\end{minipage}\hspace{0.03\textwidth}%
\begin{minipage}[b]{0.50\textwidth}
\centering
\includegraphics[height=6.6cm]{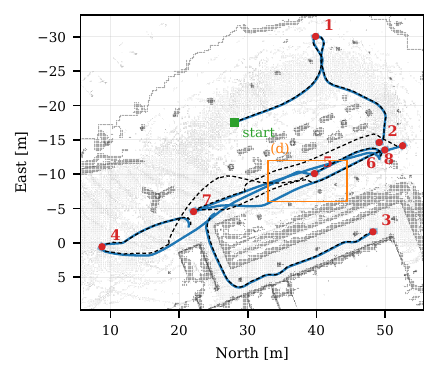}\\[2pt]
\small (b)
\end{minipage}\\[6pt]
\begin{minipage}[b]{0.40\textwidth}
\centering
\includegraphics[height=4.4cm]{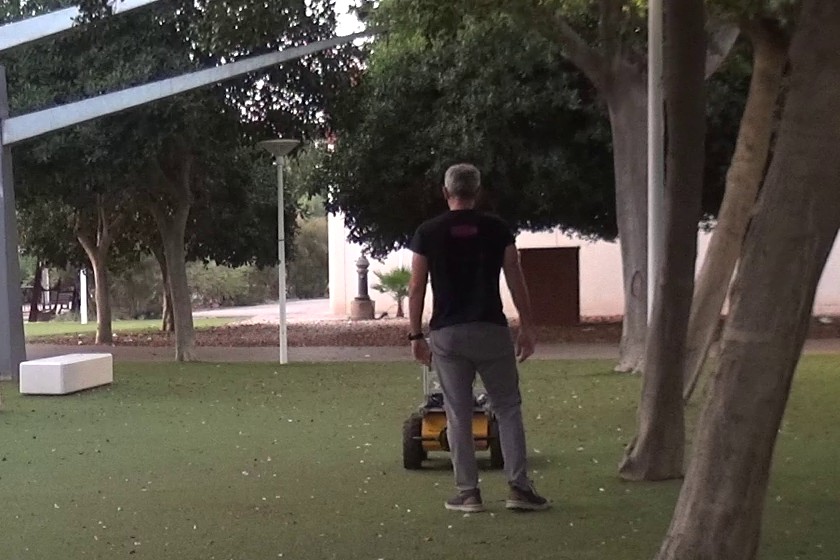}\\[2pt]
\small (c)
\end{minipage}\hspace{0.03\textwidth}%
\begin{minipage}[b]{0.50\textwidth}
\centering
\includegraphics[height=4.4cm]{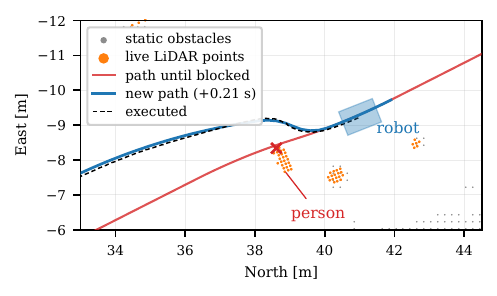}\\[2pt]
\small (d)
\end{minipage}
\caption{Real-robot experiment on the University of Almer\'{\i}a campus.
(a) The robot during a run. (b) Map (gray), static obstacles (black), first
path planned to each reached goal (blue), executed trajectory (dashed), goals
as in Table~\ref{tab:robot} (red), footprint at each arrival; where the dashed
line leaves the blue one (runs 7, 8), replanning replaced the first path;
orange: area of (d). (c) Run~7: a person stops on the path. (d) Its replan:
live LiDAR obstacle points (orange) of the scan in which the path (red) was
found blocked, new path (blue) sent $0.21$\,s later, footprint at that moment,
and executed trajectory (dashed).}
\label{fig:robot}
\end{figure*}

We deployed TPS-A* on AGRICOBIOT~I, a Clearpath Husky (skid-steer, $1.04
\times 0.72$\,m footprint) adapted for transport tasks alongside workers in
greenhouses~\cite{canadas2026greenseg,canadas2026ros2bench}, here
carrying two 3D LiDARs (Ouster OS0-32 and Velodyne VLP-16), in a campus park with trees, benches, a pergola, and people walking
by (Fig.~\ref{fig:robot}). Localization is done with
MOLA~\cite{blanco2025mola_lo} in a point-cloud map built beforehand, and the
whole stack runs on the onboard computer (Intel Core i7-8565U), with the
planner in its ROS~2 \texttt{mrpt\_navigation} wrapper (library commit
\texttt{520c770}, which includes the benchmarked one). It uses forward and
reverse C-PTGs ($v_\text{max} = 1$\,m/s, $\omega_\text{max} = 60^\circ$/s) and
an in-place rotation PTG, a $0.30$\,m and $15^\circ$ lattice, $\varepsilon =
2$, the Reeds-Shepp heuristic and shots, and a clearance cost that vanishes at
$0.8$\,m. Obstacles are points, the native input of the TP-Space query: a
static 2D layer of the map, without the objects that moved while it was built,
plus the points of the latest scans of both LiDARs that the robot body could
hit, so the live obstacles need no rasterization.

The planner runs continuously: every second, the search root moves to the
pose $3$\,s ahead of the robot on its path, the path up to there is committed,
and the search grows the same tree with the latest obstacles until it reaches
the goal. In each cycle the rest of the path is checked against the current
obstacles; a blocked path is cut before the obstacle and searched again, from
the robot if the obstacle lies before the committed end, starting where the
robot will be $3$\,s later so that it need not stop. Without a full path after
$2$\,s, the best partial one is sent while the search goes on. A trajectory
follower tracks the path, with a collision guard that slows or stops the robot
if its current motion would hit a live obstacle.

\begin{table}[t]
\centering
\caption{Real-robot runs: the eight goals reached. Paths: sent to the
follower; upd.: committed-path updates; first: from the goal to its first path;
clear.: minimum footprint distance to the static obstacles along the executed
trajectory; goal error: final position and heading.}
\label{tab:robot}
\footnotesize
\setlength{\tabcolsep}{2.5pt}
\begin{tabular}{lccccccc}
\toprule
Run & Dist. & Duration & Paths & Upd. & First & Clear. & Goal error \\
    & [m]   & [s]      &       &      & [s]   & [m]    & [cm] / [$^\circ$] \\
\midrule
1 & 21.9 & 45.7 & 1 & 10 & 0.4 & 1.02 & 6.7 / 2.0 \\
2 & 22.7 & 45.9 & 1 & 16 & 0.7 & 0.45 & 6.3 / 10.4 \\
3 & 33.7 & 68.1 & 1 & 26 & 0.4 & 0.28 & 12.2 / 10.6 \\
4 & 15.3 & 30.4 & 1 &  7 & 0.2 & 1.39 & 7.6 / 5.1 \\
5 & 38.6 & 87.0 & 1 & 15 & 0.2 & 0.38 & 12.5 / 6.6 \\
6 & 12.4 & 26.6 & 1 &  7 & 0.6 & 0.74 & 5.4 / 1.1 \\
7 & 30.4 & 61.8 & 3 & 14 & 0.2 & 0.58 & 12.6 / 7.6 \\
8 & 34.7 & 74.0 & 3 & 17 & 0.6 & 0.63 & 5.6 / 3.5 \\
\midrule
Total / median & 209.6 & 439.4 & 12 & 112 & 0.4 & 0.60 & 7.2 / 5.8 \\
\bottomrule
\end{tabular}
\end{table}

An operator gave $15$ goals in a row from a web interface: the robot reached
$8$ (Table~\ref{tab:robot}; $210$\,m at a median $0.49$\,m/s), $4$ were replaced
and $2$ canceled before arrival, and one was rejected, its footprint in
collision. For the reached goals, the first path came $0.2$ to $0.7$\,s after
the goal, including the ROS~2 messaging; of the $14$ accepted goals, $12$ got
a full path within $1.0$\,s, and the other two, both canceled later, a partial
path after $2.4$\,s. These times are longer than on BARN, with paths up to
five times longer ($12$ to $35$\,m, against about $7$\,m), the clearance cost, which multiplies the
expansions (Sec.~\ref{sec:exp-barn}), and a slower computer. The committed
path was then updated about every $4$\,s. The footprint kept at
least $0.28$\,m from the static obstacles ($0.60$\,m in median over the runs),
and the robot stopped a median $7.2$\,cm and $5.8^\circ$ from the goal pose,
within the follower tolerances; the follower's collision guard, not the
planner, briefly stopped or slowed it five times.

Fig.~\ref{fig:robot}(c,d) shows the replan of run~7. A person stopped on the
path $2.2$\,m ahead of the robot, before the committed end, so the planner cut
the path and searched again from $1.0$\,m ahead of the robot, which kept moving
at $0.57$\,m/s. The new path, around the person, reached the follower $0.21$\,s
after the detection, and the robot only slowed down for a moment. In run~8, a
person blocked the path $4.7$\,m ahead, beyond the committed end, and the new
path from there arrived after $0.28$\,s.

\section{Discussion and Limitations}\label{sec:limitations}

\textbf{Contribution in context.} TPS-A* combines known ingredients (PTGs,
TP-Space, lattice search with representatives, weighted A*). What it adds is a
change in how search and collision checking interact: one certified query per
node answers the collision question for every trajectory the search may take
from it, independently of the footprint and the branching factor
(Fig.~\ref{fig:micro}(b)), and certification is what makes that reuse safe. On
BARN, TPS-A* is at least as fast as every baseline in median and the only fast
planner without colliding paths, though Nav2 has a lighter tail with pose goals;
in parking and docking, a smoother-free Nav2 matches it in success and safety,
and the advantage reduces to the guarantee by construction.

\textbf{On a real robot.} The campus runs add live obstacles, replanning while
the robot moves, and an onboard computer. The LiDAR points enter each search
directly, with no cost map to update for collision checking, and planning took
tenths of a second, within the one-second replanning period for all but two
goals.

\textbf{Limitations.} (i) Guarantees are relative to the lattice abstraction:
like Hybrid-A*, TPS-A* is not resolution complete in the strict sense, and the
suboptimality bound covers full-pose goals only with the Reeds-Shepp heuristic,
with an additive goal-cell term that is large for cars (Sec.~\ref{sec:theory});
terminating in the goal cell also leaves a final pose error of up to one cell,
unless the path ends with a Reeds-Shepp shot.
(ii) Soundness costs a margin of up to $m + r + \sqrt{2}\Delta_c$; a coarse grid
loses the tightest passages ($96.0\%$ vs.\ $100\%$ on BARN for $0.10$ vs.\
$0.05$\,m), and a finer one costs a longer offline build and a larger grid,
both paid once per footprint. (iii) The collision check is linear in the number of local
obstacle points, so for very dense clouds a rasterized check becomes cheaper per
point; certification of region obstacles also increases the point count through
boundary sampling, and a car-sized reach puts about seven times more points in
each query than on BARN. (iv) TPS-A* has a heavier tail than Nav2 (on BARN with
pose goals, mean $24$ vs.\ $10$\,ms, maximum $0.66$\,s), its paths are
$2$--$3\%$ longer, and an infeasible query ends only when the search is
exhausted, after several seconds. (v) The evaluated PTGs have fixed speed,
so we claim kinematic, not kinodynamic, feasibility; the benchmarks are
planning-only in simulation, and the real-robot runs
(Sec.~\ref{sec:exp-robot}) are a demonstration, not a statistical evaluation. (vi) The guarantee covers the planned motion against
the given obstacles, for the footprint grown by $m$; tracking and localization
errors are left to $m$ and to the rest of the stack, and moving obstacles are
not predicted: on the robot, replanning and the follower's collision guard
handled the people walking by.

\section{Conclusions}\label{sec:conclusions}
We presented TPS-A*, a deterministic lattice planner built on certified
collision checking in trajectory parameter space: one query per node returns a
provably conservative free distance along every candidate trajectory, at a cost
independent of the footprint complexity and of the branching factor, and
certified Reeds-Shepp shots reach pose and position goals exactly. No path it returned
collided under an exact, independent test. On BARN it solves every feasible
query in $2.9$\,ms in median, on par with Hybrid-A*, $45\%$ of whose paths
overlap obstacles, and $58\times$ faster than a state-lattice planner, the only
other one without colliding paths; with BARN's position goals it is $89\times$
faster than Hybrid-A*. It removes the feasibility-versus-safety tradeoff of
circular models on real floor plans, matches Hybrid-A* in automated parking,
and docks a non-convex fork footprint that no convex model admits; on a real
Husky robot, it navigated $210$\,m on a university campus among people, with
each first path in under a second and a path blocked by a person replaced
in about $0.2$\,s. Future work includes larger real-vehicle campaigns,
velocity-dependent PTGs with acceleration-aware costs, and a cheaper collision
check for large region obstacles.

\bibliographystyle{IEEEtran}
\bibliography{tps-astar-refs,myReferences,Navigation}

\end{document}